%% file: egpaper_for_review.tex
\documentclass[10pt,twocolumn,letterpaper]{article}

\usepackage{iccv}
\usepackage{times}
\usepackage{epsfig}
\usepackage{graphicx}
\usepackage{amsmath}
\usepackage{amssymb}
\usepackage{bm}
\usepackage{tabularx}
\usepackage{multirow}

\usepackage[pagebackref=true,breaklinks=true,letterpaper=true,colorlinks,bookmarks=false]{hyperref}

\iccvfinalcopy 

\def\iccvPaperID{428} 

\ificcvfinal\fi

\graphicspath{{./figures/}}

\begin{document}

\title{Scene Graph Generation from Objects, Phrases and Region Captions}

\author{Yikang Li$^{1}$, Wanli Ouyang$^{1,2}$, Bolei Zhou$^{3}$, Kun Wang$^{1}$, Xiaogang Wang$^{1}$\\
	$^{1}$The Chinese University of Hong Kong, Hong Kong SAR, China\\ 
	$^{2}$University of Sydney,  Australia \ \ \ \ \  $^{3}$Massachusetts Institute of Technology, USA 
}


\maketitle


\begin{abstract}
Object detection, scene graph generation and region captioning, which are three scene understanding tasks at different semantic levels, are tied together: scene graphs are generated on top of objects detected in an image with their pairwise relationship predicted, while region captioning gives a language description of the objects, their attributes, relations and other context information. In this work, to leverage the mutual connections across semantic levels, we propose a novel neural network model, termed as Multi-level Scene Description Network~(denoted as MSDN), to solve the three vision tasks jointly in an end-to-end manner. Object, phrase, and caption regions are first aligned with a dynamic graph based on their spatial and semantic connections. Then a feature refining structure is used to pass messages across the three levels of semantic tasks through the graph. We benchmark the learned model on three tasks, and show the joint learning across three tasks with our proposed method can bring mutual improvements over previous models. Particularly, on the scene graph generation task, our proposed method outperforms the state-of-art method with more than 3\% margin. Code has been made publicly available\footnote{\url{https://github.com/yikang-li/MSDN}}.
\end{abstract}

\input{introduction.tex}

\begin{figure*}[t]
	\begin{center}
		\includegraphics[width=\linewidth]{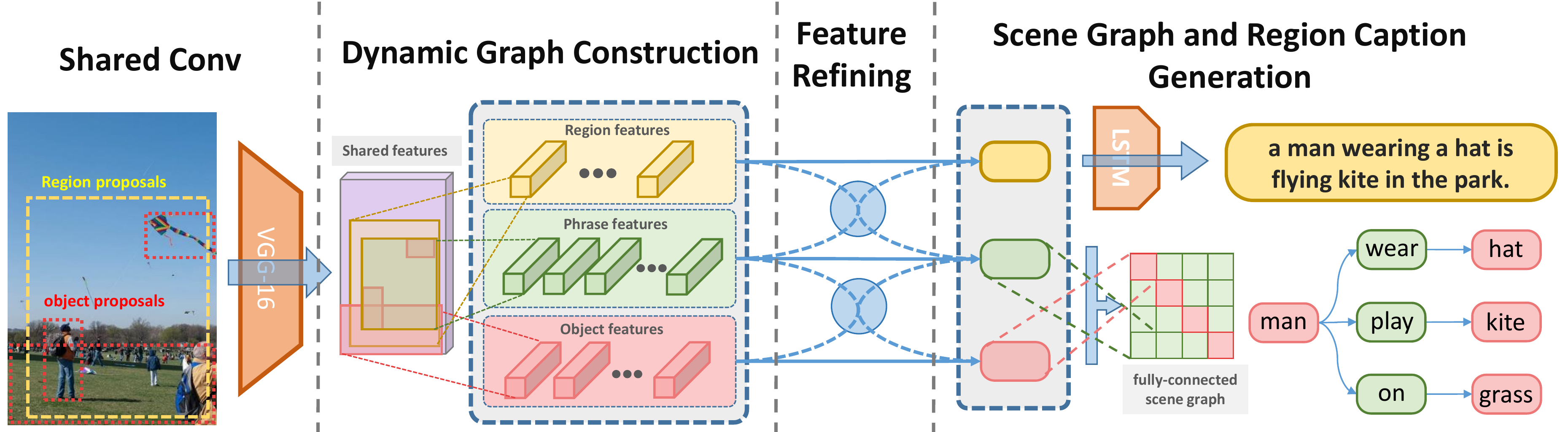}
	\end{center}
	\caption{Overview of MSDN. The two RPNs~\cite{faster_rcnn} for object and caption regions are omitted for simplicity, which share the convolutional layers with other parts. Phrase regions are generated by grouping object regions into pairs. With the region proposals for objects, phrases, and captions, ROI-pooling is used for obtaining their features. These features go through two fully connected layers and then pass messages to each other. After message passing, features for objects are used for object detection, similarly for phrase detection and region captioning. Message passing is guided by the dynamic graph constructed from the object and caption region proposals.  Features, bounding boxes and predicted labels for object~(red), phrase~(green) and region~(yellow) are assigned with different colors.}
	\label{fig:hdn}
\end{figure*}

\input{related_works.tex}

\input{model_architecture_v2.tex}

\input{experiments.tex}

\input{conclusion.tex}

\section*{Acknowledgment}
This work is supported by Hong Kong Ph.D.  Fellowship Scheme, SenseTime Group Limited, the General Research Fund sponsored by the Research Grants Council of Hong Kong (Project Nos. CUHK14213616, CUHK14206114, CUHK14205615, CUHK419412, CUHK14203015, CUHK14207814, and CUHK14239816), the Hong Kong Innovation and Technology Support Programme (No.ITS/121/15FX), National Natural Science Foundation of China (No. 61371192), and ONR N00014-15-1-2356. We also thank Xiao Tong, Kang Kang, Hongyang Li, Yantao Shen, and Danfei Xu for helpful discussions.

{\small
\bibliographystyle{ieee}
\bibliography{egbib}
}

\end{document}

%% file: introduction.tex


\section{Introduction}

Understanding visual scenes is one of the primal goals of computer vision. Visual scene understanding includes numerous vision tasks at several semantic levels, including detecting and recognizing objects, estimating the pairwise visual relations of the detected objects, and describing image regions with free-form sentences. In recent years, great progress has been made to build intelligent visual recognition systems for the three vision tasks, object detection~\cite{YOLO,faster_rcnn,SSD}, scene graph generation~\cite{visual_relationship, xu2017scene,li2017vip, Zhang_2017_ICCV}, and image/region captioning~\cite{show_attend_tell, lstm_cv, karpathy2015deep}. 

\begin{figure}[t]
	\begin{center}
		\includegraphics[width=\linewidth]{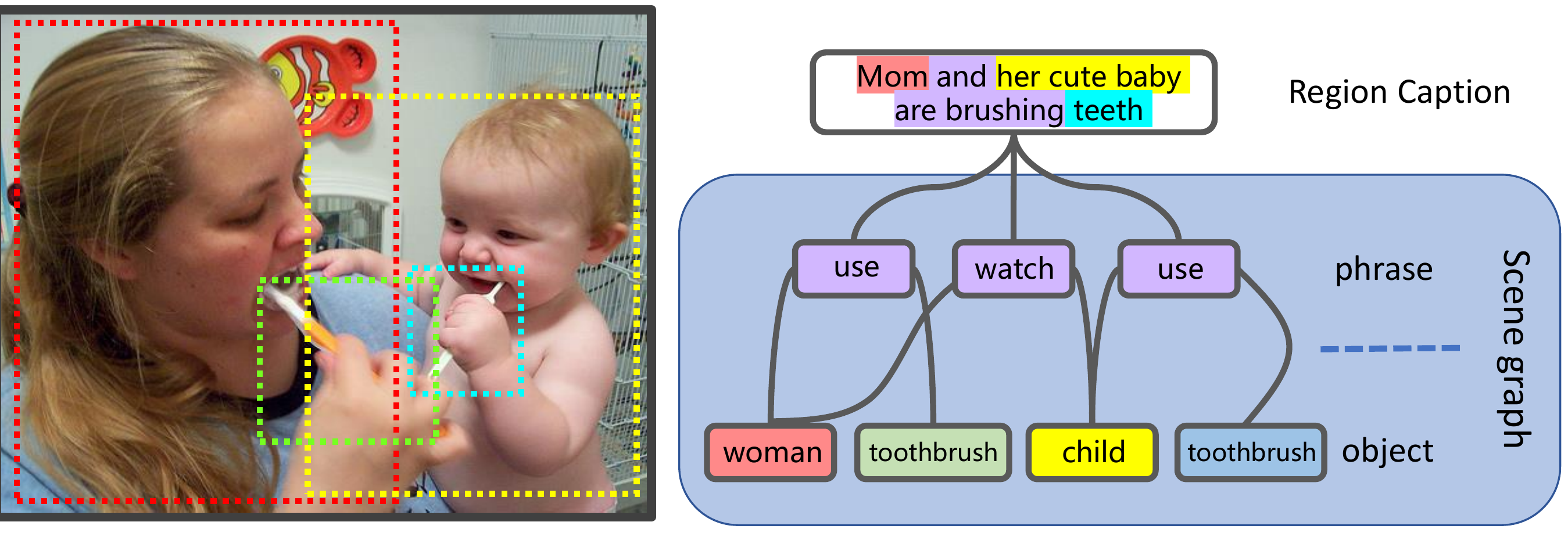}
	\end{center}
	\caption{Image with annotations of different semantic levels: objects, phrases and region captions. Scene graph is generated using all objects and their relationships in the image.}
	\label{fig:motivation}
\end{figure}

The three vision tasks target on different semantic levels of scene understanding. Take the image in Fig.\ref{fig:motivation} as an example. Object detection focuses on detecting individual objects such as woman, toothbrush, and child. Scene graph generation recognizes not only the objects but also their relationships. Such relationships can be represented by directed edges, which connect two objects as a $\langle$\emph{subject-predicate-object}$\rangle$ phrase, like $\langle$\emph{woman-use-toothbrush}$\rangle$. Region captioning generates a free-form sentence involving uncertain number of the objects, their attributes, and their interactions, as shown in Fig.\ref{fig:motivation}. We can see that though there are connections among the three tasks, the weak alignment across different tasks makes it difficult to learn a model jointly. Our work explores the possibility in understanding the image from these three levels together through a single neural network model.

The key to connect these three tasks is to leverage the spatial and semantic correlations of their visual features. For example in Fig.~\ref{fig:motivation}, the phrase $\langle$\emph{woman-watch-child}$\rangle$ provides the constraint that two persons are interacting with each other. This constraint validates the existence of the woman and child. In addition, the region caption \emph{`mom and her cute babies are brushing their teeth'} provides constraints on the  existence of the objects~(\emph{woman}, \emph{child}, and \emph{toothbrush}), their attributes~(\emph{cute}), and their relationships~(the woman is \emph{watching} the child and they are \emph{using} toothbrush) within this area. Therefore, the features for these three tasks are highly correlated and can be the complementary information of each other.  Based on this observation, we propose to jointly refine the features of different semantic levels by introducing a novel framework to align the three tasks and a message passing structure to leverage the complementary effects for mutual improvements.

In this work, we propose an end-to-end Multi-level Scene Description Network~(MSDN) to simultaneously detect objects, recognize their relationships and predict captions at salient image regions. This model effectively leverages the rich annotations at three semantic levels and their connections for image understanding. 

The contributions of this paper are summarized as follows: 1) We propose a novel model to learn features of different semantic levels by simultaneously solving three vision tasks, object detection, scene graph generation and region captioning. 2) In the model, given an image, a graph is built to align the object, phrase, and caption regions within an image.  Since  images have different objects, phrases and captions, constructed graphs could be different for different images. We provide a dynamic graph construction layer in the CNN to construct such a graph. 3) A feature refining structure is used to pass message from different semantic levels through the graph. In this way, the three tasks are integrated into one single model, and the features of three semantic levels are jointly optimized. On the Visual Genome dataset~\cite{visual_genome}, our proposed model outperforms the state-of-art methods on scene graph generation by 3.63\%$\sim$4.31\%. The mutual improvement effects are also shown on the object detection and region captioning tasks. Code has been made publicly available to facilitate further research.

%% file: related_works.tex
\section{Related Work}

\textbf{Object Detection}: 
Object detection is the foundation of image understanding. Objects serve as bricks to build up the house of the scene graph. Since CNNs were firstly introduced to the object detection by Girshick~\etal in R-CNN~\cite{rcnn}, many region-based object detection algorithms, such as Fast R-CNN~\cite{fast_rcnn}, SPP-Net~\cite{spp_net}, Faster R-CNN~\cite{faster_rcnn}, were proposed to improve the accuracy and speed. Although YOLO~\cite{YOLO} and SSD~\cite{SSD} further sped up the detection process by sharing more layers between region proposal and region recognition, Faster R-CNN~\cite{faster_rcnn} is still a popular choice for object detection because of its excellent performance. Therefore, we will adopt the pipeline of Faster R-CNN as the basis of our proposed model.

\textbf{Visual Relationship Detection}: 
Visual Relationship detection is not a new concept. It has been investigated by numerous studies in the last decade. In the early days, most works targeted specific types of phrases~\cite{choi2013understanding, desai2012detecting} or used visual phrases to improve other tasks~\cite{visual_phrase, gupta2008beyond, kumar2010efficiently, russell2006using}. Recently, researchers pay more attention to general visual relationship detection~\cite{li2017vip, xu2017scene, plummer2016phrase, zhang2017visual, dai2017detecting, zhuang2017towards, zhuang2017care} . Lu~\etal utilized the language prior in detecting visual phrases and their components in \cite{visual_relationship}. Li~\etal used the message passing structure among subject, object and predicate branches to model their dependencies~\cite{li2017vip}. Xu~\etal built up a fully-connected graph to iteratively pass messages along the scene graph~\cite{xu2017scene}. Liang~\etal applied the reinforcement learning method to the relationship and attribute detection~\cite{liang2017deep}. However, connections between phrases and captions are not built up in existing works. In this paper, we will view the objects, phrases and region captions as different semantic levels and build up their connections based on their spatial and semantic relationships. 

\textbf{Image Captioning}: 
Recently, increasingly more researchers put their attentions on interactions bwtween vision and language~\cite{li2017person, zhou2015simple, antol2015vqa, yumulti, dai2017towards}, of which, image captioning is a fantastic research topic that connects the two areas. It has been investigated for years~\cite{barnard2003matching, farhadi2010every, jia2011learning, kulkarni2013babytalk, kuznetsova2013generalizing, socher2010connecting}.  Recently, CNN plus RNN has been adopted as the standard pipeline for image captioning task~\cite{chen2014learning, donahue2015long, fang2015captions, karpathy2015deep,show_attend_tell}. Captioning was based on the whole image until the work of Johnson \etal~\cite{densecap} introduced the dense captioning task which focuses on the regions. Existing works on image/region captioning, however, do not explicitly leverage the scene graph.  Our proposed model integrates the highly-structured scene graph into our model to learn better feature for region captioning. And in return, the captioning task can also provide additional information for scene graph generation. 

\textbf{Multi-task Learning}: Multi-task learning~\cite{zhang2012convex, xue2007multi, zhang2014facial, kang2011learning, zhang2010probabilistic} has been used to model the relationships among correlated tasks. In~\cite{zhang2012convex}, a convex formulation was derived for multi-task learning. A group of related tasks was identified using statistical models in \cite{xue2007multi, zhang2014facial}.  Multi-task deep learning is used for learning facial key point detection aided by face attributes~\cite{zhang2014facial}.  Group sparsity is used in \cite{kang2011learning} to determine a group of tasks that will share feature representations. Our work propose a novel way to leverage the complementary effects from three tasks of different semantic levels.

%% file: model_architecture_v2.tex
\section{Multi-level Scene Description Network}

An overview of our proposed MSDN is shown in Figure~\ref{fig:hdn}. It adopts the region-based detection pipeline in \cite{faster_rcnn}. The model contains three parallel branches for three different vision tasks. MSDN is based on the convolutional layers of VGG-16~\cite{simonyan2014very}, which is shared by the region proposal network~(RPN) and recognition network.

The entire process can be summarized as below:
1)~Region proposal. To generate ROIs for objects, phases and, region captions. 
2)~Feature specialization. Given ROIs, to obtain specialized features that will be used for different semantic tasks.
3)~Dynamic graph construction. Dynamically construct a graph to model the connections among feature nodes of different branches based on the semantic and spatial relationships of corresponding ROIs.
4)~Feature refining. To jointly refine the features for different tasks by passing messages of different semantic levels along the graph.
5)~Final prediction. Using the refined features to classify objects, predicates and generate captions. The scene graph is generated from detected objects and their recognized relationships~(predicate).

\subsection{Region Proposal}
Three sets of proposals are generated:

\begin{itemize}
	\item object region proposals: directly generated using Region Proposal Network~(RPN) proposed in \cite{faster_rcnn};
	\item phrase region proposals: grouping the $N$ object proposals to $N(N-1)$ object pairs~(two identical proposals will not be grouped) which fully connects  object proposals with directed edges;
	\item caption region proposals: directly generated by another RPN trained with ground truth region bounding boxes.
\end{itemize}

RPNs for object and caption region proposals share the base convolutional layers of VGG-16~\cite{VGG}. The anchors of two RPNs are generated by clustering the logarithmic widths and heights of ground truth boxes  the training set using k-means clustering~\cite{hartigan1979algorithm}. To reduce the size of ROI sets, non-maximum suppression is used for object and caption ROIs separately.

\subsection{Feature Specialization}
Different branches correspond to different vision tasks. To make different branches learn their own features, we first feed the three sets of ROIs to ROI-pooling and then use different FC layer sets for different branches. In our implementation, we use two 1024-dim FC layers for each branch. After feature specialization, each branch has its own features for its specific task. 

\begin{figure}[h]
	\begin{center}
		\includegraphics[width=0.9\linewidth]{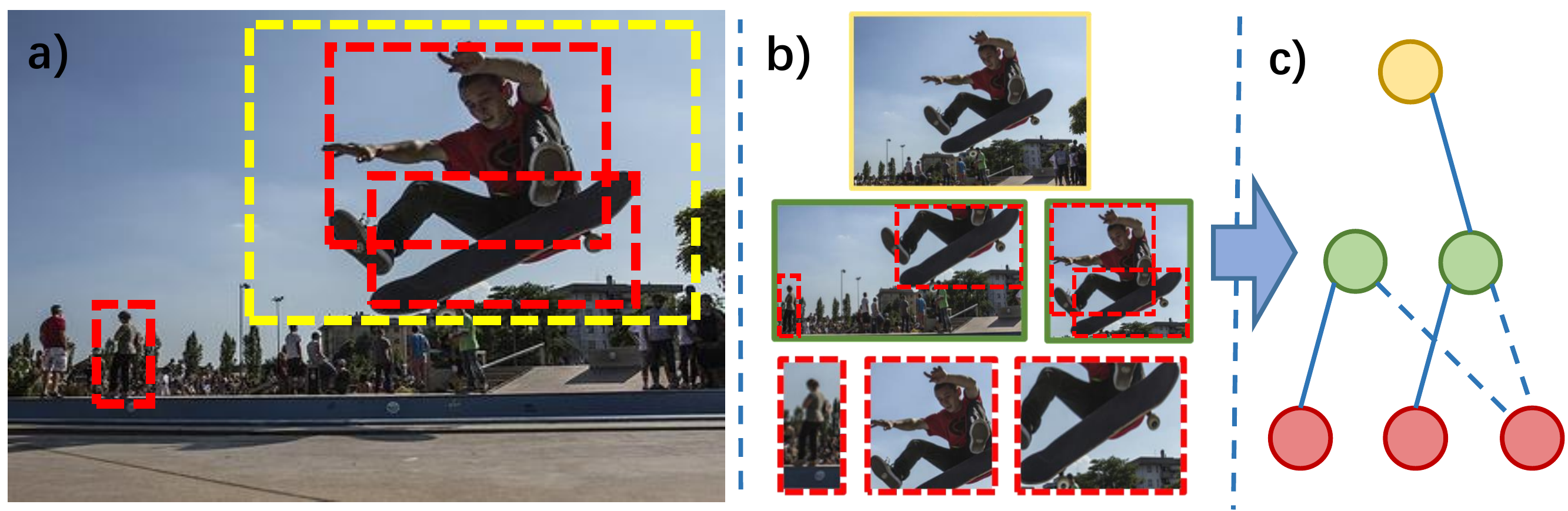}
	\end{center}
	\caption{Dynamical graph construction. (a) the input image. (b) object(bottom), phrase(middle) and caption region(top) proposals. (c) The graph modeling connections between proposals. Some of the phrase boxes are omitted.}
	\label{fig:graph_construct}
\end{figure}

\subsection{Dynamic Graph Construction}
\label{Sec:GraphLayer}
For different input images, the topology structures of the connections are different. Thus, the connection graph is dynamically built up based on the semantic and spatial relationships among the ROIs. 

Connections between phrases and objects are naturally built during constructing phrase proposals. Each phrase proposal will be connected to two object proposals as a \emph{subject-predicate-object} triplet with two directed edges. 
The connection between phrase and caption proposals is obtained based on their spatial relationship.  When a caption proposal, denoted by $\bm{b}^{(r)}$,  covers enough fraction (the threshold 0.7 is used in our experiment) of a phrase proposal, denoted by $\bm{b}^{(p)}$ , there is an undirected edge between $\bm{b}^{(r)}$ and $\bm{b}^{(p)}$. 
We ignore the direct connection between captions and objects for simplicity as they can be connected indirectly through the phrase level.

From the steps above, a graph is constructed to model the connections among objects, phrases and caption proposals. Fig. \ref{fig:graph_construct} shows an example of this graph.

The graph $\bm{G}$, contains a node set $\bm{V}$ and an edge set $\bm{E}$. For $\bm{V}$, each node in $\bm{V}$ corresponds to the specialized features of an ROI. The edge set $\bm{E}$ contains a set of the undirected edges between caption and phrase, $\bm{E}_{p,r}$, and two directed edge set, $\bm{E}_{s,p}$ and $\bm{E}_{o,p}$, where $s$ and $o$ denotes the \emph{subject} and \emph{object} in the phrase. In the following sections, we will use the denotations for simplicity.

\subsection{Feature Refining}\label{sec:feature_refining}

\begin{figure}[t]
	\begin{center}
		\includegraphics[width=\linewidth]{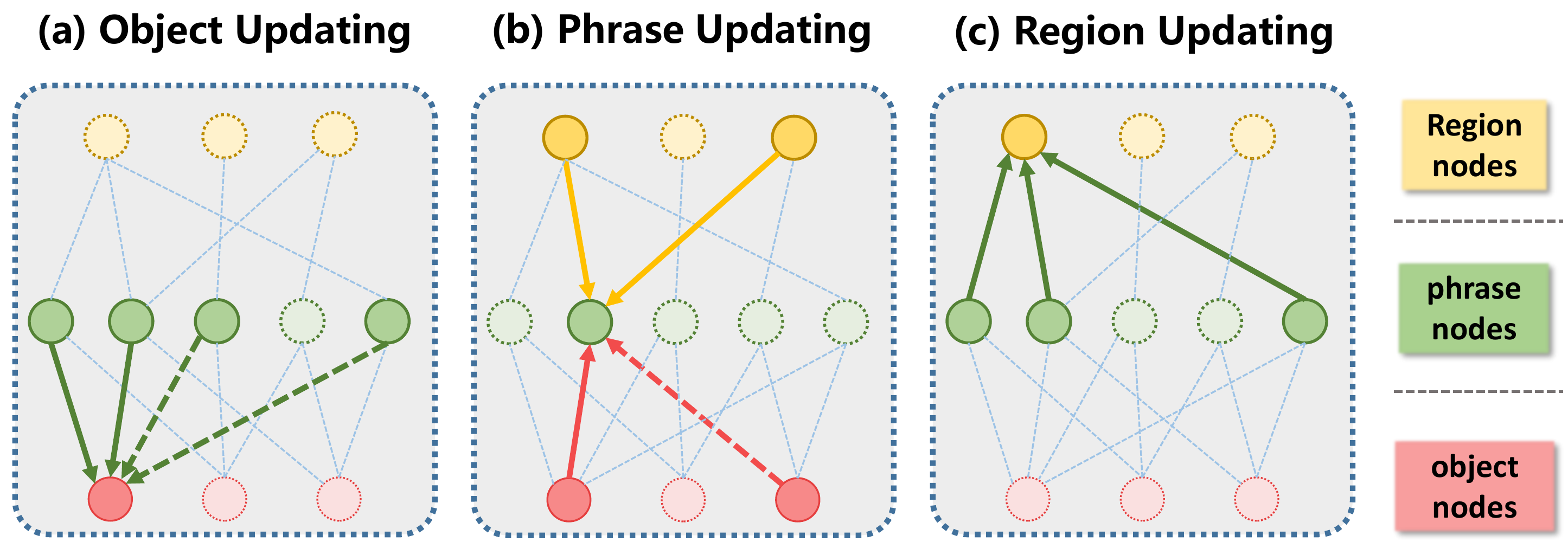}
	\end{center}
	\caption{Feature refining for object nodes \textbf{(a)}, phrase nodes \textbf{(b)} and caption nodes \textbf{(c)}. The arrow means passing direction. The two kinds of  lines connected to the object nodes are used to distinguish the \emph{subject-predicate} and \emph{predicate-object} connections. }
	\label{fig:updating}
\end{figure}

After determining the connections between different levels of nodes, message is passed among features through the edges of the graph. We divide the feature refining procedure into three parallel steps, object refining, phrase refining and caption refining~(Figure~\ref{fig:updating}). In addition, the refining procedure can be applied iteratively.

We will analyze the message passing from phrase nodes to object nodes as an example. And it can be extended to message passing between other types of nodes.

\subsubsection{Refining Features of Objects}

For each object node, there will be two kinds of connections, \emph{subject-predicate} and \emph{predicate-object}. We merge the phrase features into two sets according to the connection type and then refine the object feature with the merged phrase features. 


\textbf{Phrase feature merge}. Since the features from different phrases have different importance factors for refining objects, we use a gate function to determine weights. The features from multiple phrases are averaged by the gate as follows~(we use \emph{subject-predicate} as an example):
\begin{equation}
\tilde{\bm{x}}^{(p\rightarrow s)}_{i} = \frac{1}{\left\lVert\bm{E}_{i,p}\right\lVert} \sum_{(i, j)\in \bm{E}_{s,p}}\sigma_{\langle o,p\rangle}\left(\bm{x}^{(o)}_{i}, \bm{x}^{(p)}_{j}\right) \bm{x}^{(p)}_{j}
\label{eq:refine}
\end{equation}
where $\tilde{\bm{x}}^{(p\rightarrow s)}_{i}$ denotes the average of gated features from the phrase that connects the object by the  \emph{subject-predicate} connections with the $i$-th object node. $\bm{E}_{s,p}$ is the set of \emph{subject-predicate} connections and $\left\lVert\bm{E}_{i,p}\right\lVert$ denotes the number of phrases connected with the $i$-th object as the $\langle subject-predicate\rangle$ pairs. $\sigma_{\langle o,p\rangle}$ denotes the gate function for the object-phrase connections which is controlled by the source and target features:
\begin{equation}
\sigma_{\langle o,p \rangle}\left(\bm{x}^{(o)}_{i}, \bm{x}^{(p)}_{j}\right) = \sum_{g=1}^{G}\text{sigmoid}\left(\bm{w}^{(g)}_{\langle o,p \rangle}\cdot \left[\bm{x}^{(o)}_{i}, \bm{x}^{(p)}_{j}\right]\right),
\label{eq:gate_function}
\end{equation}
where $G$ denotes the number of the gate  templates for the input features, and we use 128 in our experiment. Each $g$ of $\bm{w}^{(g)}$ corresponds to a template. When the input feature matches the template, the value after sigmoid will be 1, and the gate will open. 
The weights $\bm{w}^{(g)}_{\langle o,p \rangle}$ are learned. 
Similar to the procedure in (\ref{eq:refine}), we can obtain the merged features $\tilde{\bm{x}}^{(p\rightarrow o)}_{i} $ for the \emph{predicate-object} connections.

\textbf{Object feature refining.}
For the $i$-th object, there are two merged features, $\tilde{\bm{x}}^{(p\rightarrow s)}_{i}$ and $\tilde{\bm{x}}^{(p\rightarrow o)}_{i}$. Then refine the $i$-th object feature as follows:
\begin{equation}
\bm{x}^{(o)}_{i, t+1} = \bm{x}^{(o)}_{i,t} + \bm{F}^{(p\rightarrow s)}\left(\tilde{\bm{x}}^{(p\rightarrow s)}_{i}\right) + \bm{F}^{(p\rightarrow o)}\left(\tilde{\bm{x}}^{(p\rightarrow o)}_{i} \right)
\label{eq:residual}
\end{equation}
where $t$ denotes the refining step since the feature refining can be done iteratively.  
$\bm{F}(\cdot) = W\cdot ReLU(\cdot)$, which is implemented by a ReLU followed by an FC layer because all the features in Eq.~\ref{eq:residual} are pre-ReLU ones. Since the merged features  $\tilde{\bm{x}}^{(p\rightarrow s)}_{i}$ and $\tilde{\bm{x}}^{(p\rightarrow  o)}_{i}$ are in the domain of phrase features, we use additional FC layers, $\bm{F}^{(p\rightarrow s)}$ and $\bm{F}^{(p\rightarrow o)}$, for modality transformation. In addition, the two FC layers do not share parameters.

\subsubsection{Refining Features of Visual Phrase and Caption}

Each phrase node is connected to two object nodes, which are $subject$ and $object$ in the $\langle subject-predicate-object\rangle$ triplet. And each caption node connects several phrase nodes. Similar to the procedure in refining features of objects, the refinement for phrase and caption also adopt the Merge-and-Refine paradigm:

\begin{equation}\label{eq:transfer_phrase}
\begin{split}
\bm{x}^{(p)}_{j, t+1} = &\ \bm{x}^{(p)}_{j,t} + \bm{F}^{(s\rightarrow p)}\left(\tilde{\bm{x}}^{(s\rightarrow p)}_{j}\right)   \\
& + \bm{F}^{(o\rightarrow p)}\left(\tilde{\bm{x}}^{(o\rightarrow p)}_{j} \right) + \bm{F}^{(r\rightarrow p)}\left(\tilde{\bm{x}}^{(r\rightarrow p)}_{j} \right), \\
\bm{x}^{(r)}_{k, t+1} = &\ \bm{x}^{(r)}_{k,t} + \bm{F}^{(p\rightarrow r)}\left(\tilde{\bm{x}}^{(p\rightarrow r)}_{k}\right) ,
\end{split}
\end{equation}
where $\bm{x}^{(p)}_{j, t+1}$ and $\bm{x}^{(r)}_{j, t+1}$ are respectively the refined phrase features and caption features at time step $t+1$. $\tilde{\bm{x}}^{(s\rightarrow p)}_{j}$ and $\tilde{\bm{x}}^{(o\rightarrow p)}_{j}$ denote the features merged from its subject and object respectively in the \emph{subject-predicate-object} phrase for $j$-th phrase node, and $\tilde{\bm{x}}^{(r\rightarrow p)}_{j}$ denotes the feature merged from its connected caption nodes. $\tilde{\bm{x}}^{(p\rightarrow r)}_{k}$ are the merged feature for the $k$-th caption node.

With this feature refining structure, messages are passed through the graph to update the features of objects, phrases, and captions by absorbing supplementary information from the connected nodes. 

\subsection{Scene Graph Generation}
Since the feature refining step has pass message between object and phrase nodes, object and corresponding pair-wise relationship categories are predicted directly based on the features of objects and phrases.
 
We use a matrix to represent the scene graph, where the element $(i,i)$ at diagonal position is the $i$th object and the element at the $(i,j)$ position for $i\neq j$ is the phrase representing the relationship between the $i$th and $j$th object. For the $i$th object, it is predicted as an object class or $\langle background\rangle$ from its refined object features. Similarly, the $(i,j)$th phrase is predicted  as a pre-defined predicate class or $\langle irrelavant\rangle$ for subject $i$ and object $j$ from phrase features. Then the scene graph is generated based on the matrix. If the object $i$ and $j$ are not classified as $\langle background\rangle$ and the predicate $(i,j)$ is not $\langle irrelavant\rangle$, then the two objects are connected through the predicate $(i,j)$. In this way, we will get a scene graph based on the matrix.

\subsection{Region Caption Generation}

Different from the object and phrase nodes, the region features contains a wide range of information, such as objects, their interactions and attributes, scene-related information, \etc. Therefore, we feed them into an LSTM-based language model to generate natural sentences to describe the region. We adopt the vanilla language model widely used for image captioning~\cite{densecap, karpathy2015deep}.

The language model takes the image vector as input and outputs a free-form sentence to describe the content in the region. The model consists of four parts: 1) an image encoder, which is used to transform the image feature to the same domain of word features; 2) a word encoder, to transform the one-hot vector to a word embedding; 3) a two-layer LSTM model, which is to encode the image information and the temporal dependencies within the sequence; 4) a word decoder, which is used to decode the output feature of LSTM to a distribution over words. 

At the first time step, image vectors are transformed to the same domain of word vectors by image encoder. Then coded image feature will be fed into a two-layer LSTM. At the second step, the $\langle start\rangle$ token will be fed into the model to indicate the start of the sentence. Then the predicted word at time $t$ will be fed into the model as input until the $\langle end\rangle$ or the maximum length is reached.

%% file: experiments.tex
\section{Experiment}
Scene graph generation can be viewed as an intermediate task connecting the object detection and region captioning, which aims at capturing the structural information of an image with a set of pair-wise relationships. Compared to object detection, the scene graph generation measures the feature learning from more aspects. And different from the region captioning, the performance of the scene graph generation model is easier to measure quantitatively and it excludes the influence brought by the different language model implementations. Therefore, the experiment part mainly focuses on this task.

Some explanatory experiments are also done on the object detection and region captioning tasks to show mutual improvements brought by the joint inference across semantic levels.

\subsection{Dataset}
\label{sec:dataset}
All the experiments are done on the Visual Genome~\cite{visual_genome} dataset. The objects and relationships are from the Relationship subset, and the region caption annotations are based on the Region Description subset. The two subsets share the image but target on different tasks.

First, we do some preprocessing on the relationship annotations. We normalize the words in different tenses and then select the top-150 frequent object categories and top-50 predicate categories. Moreover, the object boxes whose shorter edges are smaller than 16 pixels are removed. After preprocessing, there are 95998 images left.

For the remaining 95998 images, we further pre-process the region caption annotations. All the words are changed to lower case. Top-10000 frequent words~(including punctuations) are used to build up the dictionary and all the other words are changed to $\langle  unknown\rangle$ token. In addition, all the small regions with shorter edges smaller than 32 are removed. NLTK~\cite{nltk} is used to tokenize the sentence. 

After the two preprocessing steps above, a cleansed dataset containing the annotations of localized objects, phrases and region descriptions are built for our experiments. From the 95998 images in the dataset, 25000 images are sampled as the testing set and the remaining 70998 images are used as the training set.

\begin{table*}[t]
	\renewcommand{\arraystretch}{1.1}
	\setlength{\tabcolsep}{4pt}
	\small
	\begin{center}
		\begin{tabularx}{1.0\linewidth}{c| cccc | cccccc}
			\hline
			\multirow{2}{*}{ID} & \multirow{2}{*}{Message Passing} & \multirow{2}{*}{Cap. branch} & \multirow{2}{*}{Cap. Supervision} & \multirow{2}{*}{FR-iters} & \multicolumn{2}{c}{PredCls} & \multicolumn{2}{c}{PhrCls} & \multicolumn{2}{c}{SGGen} \\
			& & & & & Rec@50& Rec@100 & Rec@50 & Rec@100 & Rec@50 & Rec@100 \\
			\hline
			1 & -  & - & - & 0 & 49.28 & 52.69 & 7.31& 10.48 & 2.39 & 3.82 \\
			2 & \checkmark & - & - & 1 & 63.12 & 66.41 & 19.30 & 21.82 & 7.73 & 10.51\\
			3 &\checkmark & \checkmark & - & 1 & 63.82 &  67.23& 20.91 & 23.09 & 8.20 & 11.35 \\
			4 &\checkmark & \checkmark & \checkmark & 1 & 66.70 & \textbf{71.02} & 23.42 & 25.68 & 10.23 & 13.89\\
			5 &\checkmark & \checkmark & \checkmark & 2 & \textbf{67.03} & 71.01& \textbf{24.22} & \textbf{26.50} & \textbf{10.72} & \textbf{14.22} \\
			6 &\checkmark & \checkmark & \checkmark & 3 & 66.23 & 70.43& 23.16&  25.28& 10.01& 13.62\\
			\hline
		\end{tabularx}
	\end{center}
	\caption{Ablation studies of the proposed model. \textbf{PredCls} denotes predicate recognition task. \textbf{PhrCls} denotes phrase recognition task. \textbf{SGGen} denotes the scene graph generation task. \textbf{Message passing} denotes whether to add feature refining structure to pass message.  \textbf{Cap. branch} denotes whether to use the caption branch as an extra connection source. \textbf{Cap. Supervision} indicates whether to use region caption annotation as the supervision to guide the learning of the caption branch. \textbf{FR-iters} denotes the number of feature refining iterations.}
	\label{tab:component}
\end{table*}

\subsection{Implementation Details}

\textbf{Model training details} 
Our model is initialized on the ImageNet pretrained VGG-16 network~\cite{VGG}. To reduce the number of parameters, we only use 1024 neurons of the fully-connected layers from the original 4096 ones and then scale up the weights accordingly as initialization. The newly introduced parameters are randomly initialized. We first train RPNs and then jointly train the entire model from the base learning rate 0.01 using SGD with gradients clipping. The parameters of VGG convolutional layers are fixed at first, and then trained with 0.1 times the learning rates of other layers after the first decay of the base learning rate. In addition, there is no weight decay for the language model and the parameters are updated using Adam. 

\textbf{Loss Functions}
For the object branch, we use the cross-entropy loss for the object classification and the smooth L1 loss for the box regression. For the phrase branch, the cross-entropy loss is used for predicting the labels of predicates. For the caption branch, the cross-entropy loss is used for generating the every word of free-form sentences and the smooth L1 loss is used for regressing the corresponding proposals. Three losses are summed up equally. Since every step at feature refining parts is differentiable, BP can be applied for the feature refining part.

\textbf{Mini-batch preparation for training} 
A mini-batch contains one image. After generating proposals with RPN layers, we use 0.7 and 0.75 as the NMS threshold for object proposals and caption proposals respectively and keep at most 2000 boxes after NMS. Then we sample 256 object proposals and 128 caption proposals from each image. As the number of phrase proposals is too large and the positive samples are sparse, we sample 512 with 25\% positive instances. In addition, we assign $\langle irrelavant\rangle$ to the negative phrase samples, $\langle background\rangle$ to the negative objects, and the $\langle end \rangle$ to the negative caption proposals.

\textbf{Details for inference.} In testing, we set the NMS threshold to 0.35 and 0.45 for object and caption region proposals. After the graph for the image is constructed,  features from all the sampled proposals are used for refining their features.

\subsection{Evaluation on Scene Graph Generation}
\subsubsection{Experiment settings}
\label{sec:exp_setting}

\textbf{Performance Metric.} Following \cite{visual_relationship}, the \emph{Top-K recall}~(denoted as \emph{Rec@K}) is used as the main performance metric, which is the fraction of the ground truth instances hit in the top-$K$ predictions. The reason of using \emph{recall} instead of \emph{mean average precision(mAP)} is that the annotations of the relationships are incomplete. \emph{mAP} will falsely penalize the positive but unlabeled predictions. In our experiment,  \emph{Rec@50} and \emph{Rec@100} will be reported. 

\textbf{Task Settings.} Since scene graph generation involves the classification of the $\langle$\emph{subject-predicate-object}$\rangle$  triplet and localization of objects. We evaluate our proposed model on three sub-tasks of scene graph generationz proposed in \cite{xu2017scene}:
\begin{itemize}
	\item \textbf{Predicate Recognition} (PredCls): To recognize the relationship between the objects given the ground truth location of object bounding boxes. This task is aimed at examining the model performance on the classification of the predicates alone.

	\item \textbf{Phrase Recognition} (PhrCls): To predict the predicate categories as well as the object categories given the ground-truth location of objects. This task evaluates the model performance on the recognition of both predicates and objects.

	\item \textbf{Scene Graph Generation} (SGGen): To detect objects and recognize their pair-wise relationships. The object is correctly detected if it is correctly classified and its overlap with the ground truth bounding box is larger than 0.5. A relationship is correctly detected if both the $subject$ and $object$ are correctly detected and the $predicate$ is correctly predicted. The location of objects is not provided.
\end{itemize}

\subsubsection{Comparison with existing works}

We compare our proposed MSDN with the following methods under the three task settings:  (1) The model using Language Prior (\textbf{LP})~\cite{visual_relationship}, which detects objects first and then estimate the categories of predicate using visual features and word embeddings. (2) Iterative Scene Graph Generation~(\textbf{ISGG})~\cite{xu2017scene}, which uses the iterative message passing along the scene graph with a GRU-based feature refining scheme. We have reimplemented their model. The model is trained and tested on the cleansed dataset mentioned in Section \ref{sec:dataset}.  All the methods are based on the VGG-16 model.

\begin{table}[h]
	\small
	\begin{center}
		\begin{tabular}{l  l ||  c c c  }
			\multicolumn{2}{c}{Task} & LP~\cite{visual_relationship}  & ISGG~\cite{xu2017scene}  & Ours  \\ \hline\hline
			\multirow{2}{*}{PredCls}     &R@50  & 26.67  & 58.17       &\textbf{67.03} \\
			&R@100  & 33.32 & 62.74        &  \textbf{71.01}      \\\hline
			\multirow{2}{*}{PhrCls}      &R@50& 10.11   & 18.77       &  \textbf{24.34}      \\
			&R@100 &  12.64  & 20.23      &  \textbf{26.50}\\\hline
			\multirow{2}{*}{SGGen}      &R@50& 0.08    & 7.09        &  \textbf{10.72}     \\
			&R@100 & 0.14    & 9.91      &  \textbf{14.22}      \\
			\hline
		\end{tabular}
	\end{center}
	\caption{Evaluation on the Visual Genome
		dataset~\cite{visual_genome}. We compare our proposed model with existing works on the three tasks illustrated in Sec.~\ref{sec:exp_setting}. The result of LP is reported in~\cite{xu2017scene}. ISGG is reimplemented by ourselves and evaluated on our cleansed dataset. }
	\label{table:vg_eval}
\end{table}

From the results in Table~\ref{table:vg_eval}, we can see that our proposed model performs better than the existing works. Compared to the ISGG model~\cite{xu2017scene}, our model introduces the caption branch to provide more context information for phrase recognition. In addition, our model passes message as residual, which makes the model easier to train.

\subsubsection{Component Analysis}

There are many components that influence the performance of MSDN. Table~\ref{tab:component} shows our investigation on the performance of different settings of MSDN on the Visual Genome dataset~\cite{visual_genome}.

\textbf{Message passing.} Model 1 in Table~\ref{tab:component} is the baseline that does not use message passing to refine features and does not have the branch for caption. Model 2 is based on Model 1 and passes message between related object and phrase nodes. By comparing Model 1 and 2 in Table~\ref{tab:component}, we can see that passing message with the feature refining structure  proposed in Sec.~\ref{sec:feature_refining} can help to leverage the connection between the objects and phrases, which significantly improves the model performance by $5.34\%\sim 6.69\%$ on SGGen task. 

\textbf{Caption region branch.} Based on Model 2, Model 3 only has an extra caption branch without the caption supervision. We remove the LSTM language model in Fig.~\ref{fig:hdn} and only use the caption branch as extra context information source. Model 3 has $0.47\%\sim0.84\%$ gain when compared with Model 2.  This improvement is more likely to come from the more parameters introduced by the caption branch.

\textbf{Region Caption Supervision.} Model 4 further uses additional supervision of region caption sentences for the region caption branch. It outperforms Model 3 by $2.03\%\sim 2.64\%$. The improvement mainly comes from the complementary features learned with additional information.  Supervision helps the caption branch learn it own specialized features, which can provided extra information for other branches. Compared to the object and predicate categories, region captions provide another way to understand the image. 

\textbf{The number of feature refining iterations.} Model 4$\sim$6 are different in the number of iterations in message passing. By comparing Model 4$\sim$6, the results show that two iterations may be the optimal settings for the scene graph generation. Compared to Model 4 with one iteration, Model 5 with two iterations constructs the connection between captions and objects indirectly, which brings $0.33\%\sim 0.49\%$ gain. However, more iterations make the model harder to train. Therefore, when we refine the features for three iterations, the training issue suppress the gain brought by the better feature refining. Therefore, the performance of Model 6 will deteriorate by $0.21\%\sim 0.27\%$.

\begin{figure*}[t]
	\begin{center}
		\includegraphics[width=\linewidth]{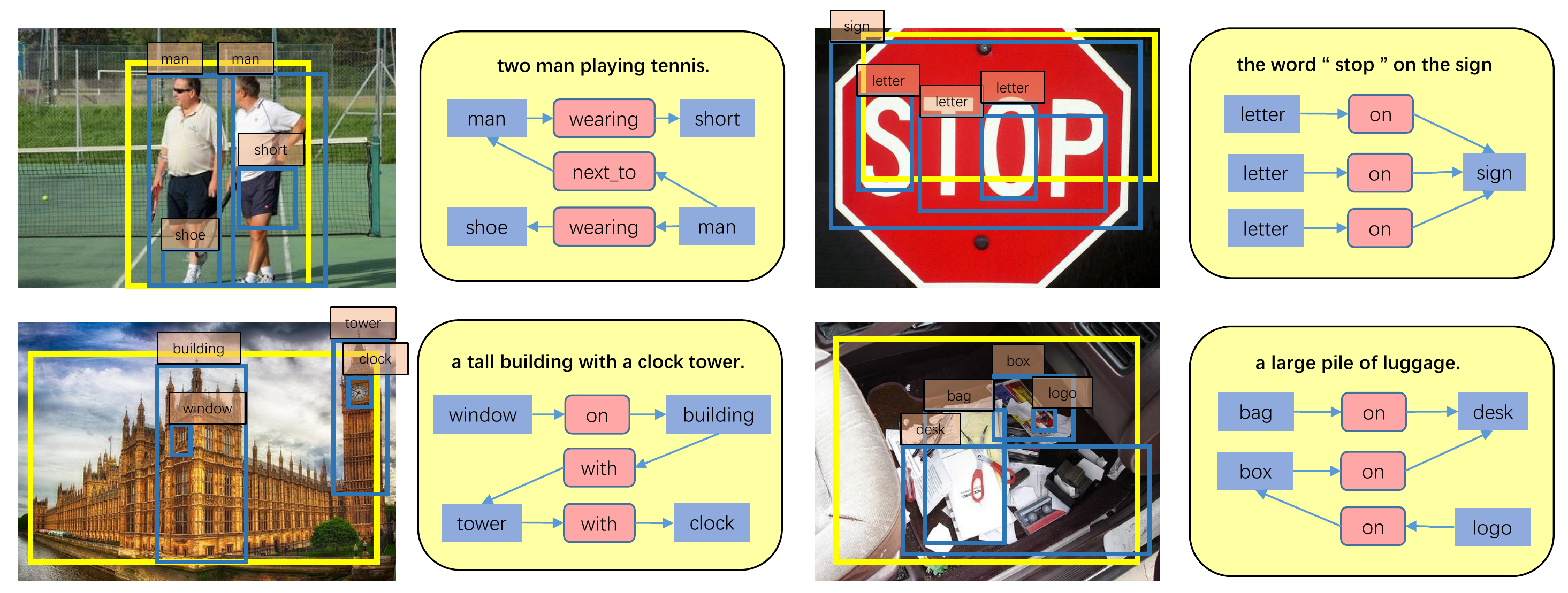}
	\end{center}
	\caption{Qualitative results for region captioning. The most salient regions with captions are shown~(yellow boxes). We also show several relationships that are connected to the captions. The connection is built by our proposed dynamic graph generation in Sec.~\ref{Sec:GraphLayer}. }
	\label{fig:result}
\end{figure*}

\subsection{Evaluation on Object Detection}\label{sec:object_detection}

We further evaluate our proposed MSDN on object detection task. 

\textbf{Setup.} We directly use the objects within the dataset prepared in \ref{sec:exp_setting}. All the objects have at least one relationship with other objects. We adopt the mean Average Precision~(mAP) metric as one evaluation metric. In addition, as most of the objects are small, poor localization of the objects highly influences the mAP metrics, we also report the accuracy of the object classification with the ground truth bounding boxes given.

We compare our proposed MSDN with Faster R-CNN~\cite{faster_rcnn}~(\textbf{FRCNN}) trained on the same dataset. In addition, to check whether the additional supervision can benefit the feature learning of convolutional layers, we also show the results for the model with the feature refining structure removed~(\textbf{Baseline-3-bran.}) and use the object branch for object detection~(like the model 1 in \ref{tab:component}). 

\begin{table}[t]
	\small
	\vspace*{-5pt}
	\begin{center}
		\begin{tabular}{l || c c c  }
			\hline 
			\textbf{Object Det.} & FRCNN~\cite{faster_rcnn} &  Baseline-3-bran. & Ours \\\hline
			mean AP(\%)    & 6.72  & 6.70  &     \textbf{7.43}   \\
			Acc. Top-1(\%)   &  53.57  &  53.14 &    \textbf{61.12}     \\
			Acc. Top-5(\%)   &  83.50 &  83.25 &  \textbf{89.86}       \\\hline\hline
			\textbf{Region Caption} & Baseline &  Baseline-3-bran. & Ours \\\hline
			AP~\cite{densecap}(\%)   & 4.41 & 4.28 &  \textbf{5.39}       \\
			\hline
		\end{tabular}
		\vspace*{-3pt}
	\end{center}
	\caption{Object detection and region captioning results evaluated on Visual Genome
		dataset~\cite{visual_genome}. \emph{Baseline-3-bran.} has 3 separate branches without message passing.}
	\label{table:other_task}
\end{table}

\textbf{Results.}  Since the Visual Genome Dataset has many object classes that are small and hard to detect, the mAP is small for all approaches. Nevertheless, our model outperforms Faster R-CNN and baseline model with three separated branches on the Visual Genome Dataset, because our model introduces more context information from phrases and captions~( when trained with more than two iterations) to the objects as complementary source, which serves as visual cues to help recognize objects.

\subsection{Evaluation on Region Captioning}
We further evaluate our model on the region caption task. 

\textbf{Setup.}  We adopt the evaluation metric proposed by Johnson~\etal in \cite{densecap} for region captioning. It measures the mean Average Precision across a range of thresholds for both localization and language accuracy. The Meteor scores~\cite{banerjee2005meteor} are used as the language metric, because it is highly correlated with human judgments. During the evaluation, the ground truth bounding regions are merged as one region with several reference annotations if they are heavily overlapped with each other~(based on IOU with threshold of 0.7).

To make the model comparable, we re-implement the main part of Densecap~\cite{densecap} using Faster R-CNN~\cite{faster_rcnn} pipeline based on VGG-Net~(\textbf{Baseline}) and use the same language model as our proposed model. Our implementation performs comparably with the original Densecap under same settings~(4.41\% vs 4.62\% evaluated on our cleansed dataset). In addition, similar to \ref{sec:object_detection}, we also include another baseline model with three separated branch without message passing~(\textbf{Baseline-3-bran.}).  All the models are evaluated on our cleansed dataset. 

\textbf{Quantitative Results.}  
From Table.~\ref{table:other_task}, we can see that, our proposed model outperforms the other two baseline models. Because we have excluded the influence brought by the number of parameters and utilized the same language model for them, the gain is obtained by the extra information introduced through the message passing. And the messages passed to the region come from the scene graph composed by the objects and their relationships. Such structural information can help the region branch infer the content of the region. In addition, by comparing the two baseline models, simply introducing extra supervision will not improve the accuracy.  

\textbf{Qualitative Results.} 
Region captioning results with the highest score are shown in Figure~\ref{fig:result}. We also show the objects and their relationships that are connected to the captions through the dynamic graph. We can see that the region captioning result is highly correlated to the scene graph. We also observe failure case (bottom right in Figure~\ref{fig:result}, where the misclassification of objects and relationships would mislead the caption branch to recognize the region as \emph{a large pile of luggage}.

%% file: conclusion.tex
\section{Conclusion}

This paper targets on scene understanding by jointly modeling three vision tasks, \ie object detection, visual relationship detection and region captioning, with a single deep neural network in an end-to-end manner. The three tasks  at different semantic levels are tightly connected.  A Multi-level Scene Description Network~(MSDN) model is proposed to leverage such connection for  better understanding image. 
In MSDN, given an input image, a graph is dynamically constructed to establish the links among regions with different semantic meaning. The graph provides a novel way to align features from different tasks. Experimental results show that this joint inference process brings improvement in all the three tasks.